\documentclass{article}

    \PassOptionsToPackage{numbers, compress}{natbib}

    \usepackage[final]{neurips_2024}

\usepackage[utf8]{inputenc} 
\usepackage[T1]{fontenc}    
\usepackage{hyperref}       
\usepackage{url}            
\usepackage{booktabs}       
\usepackage{amsfonts}       
\usepackage{nicefrac}       
\usepackage{microtype}      
\usepackage[table] {xcolor}         
\usepackage{multirow}
\usepackage{booktabs}

\usepackage{caption}
\usepackage{adjustbox}
\usepackage{tcolorbox}
\usepackage{framed}
\usepackage{xcolor}
\usepackage{colortbl}

\title{Benchmarking Table Comprehension in the Wild}

\author{%
  Yikang Pan\thanks{Work completed during an internship at Boson AI.} \\
  Boson AI \\
  University of Toronto\\
  \texttt{yikang@boson.ai} \\
  \And
  Yi Zhu \\
  Boson AI \\
  \texttt{yi@boson.ai} \\
  \And
  Rand Xie \\
  Boson AI \\
  \texttt{rand@boson.ai} \\
  \AND
  Yizhi Liu \\
  Boson AI \\
  \texttt{yizhiliu@boson.ai} \\
}

\begin{document}

\maketitle

\begin{abstract}

Large Language Models (LLMs), while being increasingly dominant on a myriad of knowledge-intensive activities, have only had limited success understanding lengthy table-text mixtures, such as academic papers \cite{wang_leave_2024} and financial reports \cite{islam_financebench_2023}. Recent advances of long-context LLMs have opened up new possibilities for this field. Nonetheless, we identify two roadblocks: (1) Prior benchmarks of table question answering (TableQA) have focused on isolated tables without context \cite{chen_finqa_2021}, making it hard to evaluate models in real-world scenarios. (2) Prior benchmarks have focused on some narrow skill sets of table comprehension such as table recognition \cite{zheng_global_2020}, data manipulation/calculation \cite{zhang_tablellm_2024}, table summarization \cite{parikh_totto_2020} etc., while a skilled human employs those skills collectively. In this work, we introduce TableQuest, a new benchmark designed to evaluate the holistic table comprehension capabilities of LLMs in the natural table-rich context of financial reports. We employ a rigorous data processing and filtering procedure to ensure that the question-answer pairs are logical, reasonable, and diverse. We experiment with 7 state-of-the-art models, and find that despite reasonable accuracy in locating facts, they often falter when required to execute more sophisticated reasoning or multi-step calculations. We conclude with a qualitative study of the failure modes and discuss the challenges of constructing a challenging benchmark. We make the evaluation data, judging procedure and results of this study publicly available to facilitate research in this field.

\end{abstract}

\section{Introduction}

In many fields, ranging from finance to healthcare, tables play a crucial role in organizing and interpreting vast amounts of data. They allow individuals to discern patterns, extract meaningful insights, and make informed decisions by presenting information in a concise, structured format. However, the tabular format poses visible challenges for large language models (LLMs), which are fundamentally designed to process data in sequential, text-based formats rather than navigating the complex spatial relationships inherent in tables. Despite growing interest and progress in natural language processing, LLMs still struggle with extracting, understanding and reasoning about information presented in tabular form, thereby limiting their use in industrial applications.

While numerous approaches have been proposed to enhance LLMs’ performance on tasks involving tabular data, there is a significant gap in comprehensive evaluation frameworks that capture the breadth and complexity of real-world use cases. Existing benchmarks tend to be narrowly focused, often concentrating on specific aspects like table-based question answering \cite{nan_fetaqa_2022,chen_finqa_2021,deng_tables_2024,zhao_financemath_2023,zhu_tat-qa_2021,kweon_open-wikitable_2023} or  information retrieval \cite{kostic_multi-modal_2021,huang_mixed-modality_2022,wang_leave_2024}, and fail to represent the full spectrum of challenges that professionals encounter when working with tabular data.

To address this gap, we introduce \textit{TableQuest}, a novel benchmark specifically designed to evaluate the proficiency of LLMs in understanding and reasoning with tabular data across a diverse set of scenarios. 
TableQuest is a question-answering dataset over public traded companies' annual filings which contain ample text and tables. We carefully construct the questions such that on answering the questions correctly, the models demonstrate the following capabilities:

\textbf{Extraction (Easy level).} Given the popularity of long-context retrieval benchmarks (e.g., needle-in-a-haystack \cite{kamradt}), we adapt this format by focusing on retrieving information from table cells within real-world documents, rather than from randomly injected facts. It not only enables evaluating models to distinguish similar pieces of information under an ecologically valid condition, but also requires models to extract insights of two dimensions (rows and columns) from linear inputs.

\textbf{Calculations (Medium level).} Finance reports, the essential tools to make informed financial decisions and strategize for future growth, provide the perfect ground for a wide range of numerical tasks, including data analysis, forecasting, budgeting, and performance tracking. Prior works have explored this area in a wide range of heavily-engineered systems \cite{zhao_financemath_2023,cai_internlm2_2024}, breaking down the problem into multiple steps for multiple models specializing in each. In our benchmark, we present this challenge in a monolithic fashion, pressuring-testing the models' ability to spot evidence in tables as well as conduct accurate calculations.

\textbf{Analytics (Hard level).} One step further from accurate calculations, we ask models to uncover insights from numbers they extracted or calculated and output in words. The aim is that models could digest the reports and output summaries in a query-focused manner, effectively aiding humans at document comprehension.

In our evaluation of TableQuest, we explored the performance of a wide array of both proprietary and open-source LLMs, employing an automated evaluation system to assess the quality and correctness of the models' responses. Our initial results indicate that while some models can handle straightforward data extraction tasks with reasonable accuracy, they often falter when required to execute more sophisticated reasoning or multi-step calculations. Common issues include difficulty maintaining context over longer sequences, following instructions (particularly format),  inconsistencies in applying domain-specific knowledge and reasoning, highlighting areas where current models fall short and where future improvements are necessary.

\begin{figure}
    \centering
    \includegraphics[width=1.0\linewidth]{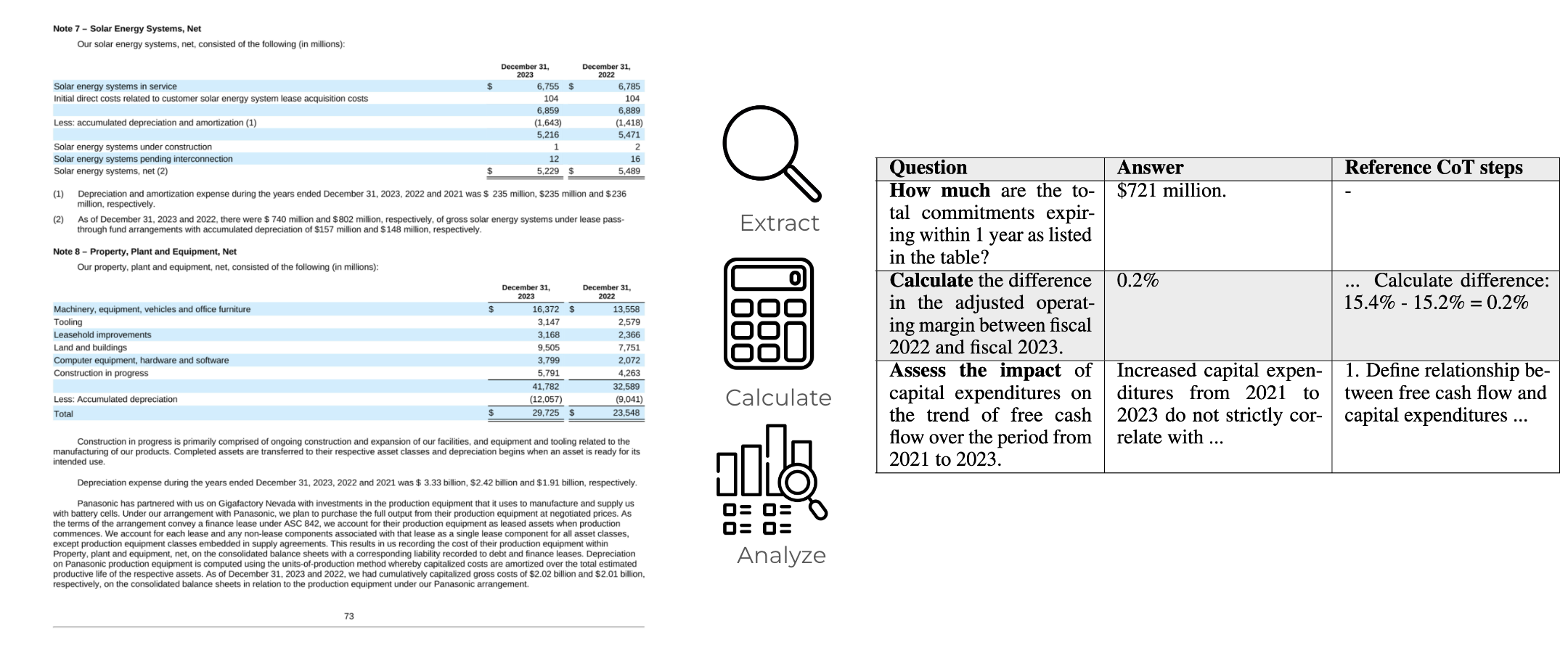}
    \caption{A sample of our TableQuest benchmark showcasing the different aspects of table comprehension. The sample image on the left was for demonstration, which was rendered from the source HTML file.}
    \label{fig:enter-label}
\end{figure}
\newpage

\section{TableQuest}

\subsection{Data Preparation} 

\textbf{Source.} We collect the 10-K reports of S\&P 500 companies directly from the SEC's EDGAR database \footnote{https://www.sec.gov/search-filings},  which offers comprehensive HTML annotations for each table, capturing hierarchical structures and detailed cell information, including text content and formatting. This structured format allows for the extraction and post-processing of data using HTML tags, enabling a thorough representation of the complex relationships and structures present within the tables. Despite a lack of consensus on the best format for table serialization \cite{fang_large_2024,lu_large_2024}, we employed HTML due to two reasons: 1. it can preserve table structures better than other languages (e.g. natural text, markdown, json etc.), and 2. the internet being a critical source for web-scale LLM pre-training corpora \cite{raffel_exploring_2023,penedo_fineweb_2024}.

\textbf{Question-Answer Synthesis.} Existing question-answering datasets often exhibit internal homogeneity, such as focusing solely on tabular answers derived from multiple tables \cite{pal_multitabqa_nodate}. To address this limitation, we implemented a progressive three-round chat format, where each round builds upon the previous one, resulting in a sequence of question-answer pairs. In each round, we introduce a new question that demands an additional capability, expanding the model's range of responses. This iterative process ensures that each prior question-answer pair informs and enhances the subsequent one, fostering greater diversity in the questions generated. Additionally, we incorporated explicit chain-of-thought (CoT) prompting to guide the model in articulating its reasoning steps, which facilitated more thorough quality inspection. We present an validation study on the quality of the question-answer generation pipeline with a self-supervised data source, which we detail in Appendix \ref{sec:appendix-f}.
The multi-round chat prompts can be found in Appendix \ref{sec:appendix-a}.

\textbf{Quality filtering.} 
Given the difficulty of questions, we employ a hybrid human-machine verification pipeline to validate the questions and answers generated were relevant, reasonable and valid. We collaborated closely with a team of finance experts to develop the initial guidelines for constructing the questions. In Appendix \ref{sec:appendix-a.2}, you can find the data processing and quality assurance specifics.

\begin{table}[h]
\centering
\begin{tabular}{|l|c|c|c|}
\hline
\textbf{Metric} & \textbf{Easy} & \textbf{Medium} & \textbf{Hard} \\
\hline
Avg. Question Length & 90.19 & 135.04 & 162.79 \\
Avg. Answer Length & 27.70 & 15.61 & 202.61 \\
\hline
\multicolumn{4}{|l|}{\textbf{Table Statistics}} \\
\hline
Avg. Tables per file & 1.82 & 1.90 & 1.86 \\
Avg. Rows per Table & 11.31 & 10.87 & 10.49 \\
Avg. Columns per Table & 13.94 & 13.32 & 13.44 \\
\hline
\end{tabular}
\vspace{5mm}
\caption{Summary Statistics for TableQuest Dataset (Eval) by Difficulty. The context for these questions are sampled randomly, such that the complexity of the tables should be on the same level.}
\label{tab:finsyn_summary_by_difficulty}
\end{table}
\vspace{-5mm}

\section{Experiments}

\textbf{Inputs and Outputs.} We sample 240 questions from our synthesized dataset, 80 of each difficulty level, as the evaluation subset of our benchmark. Each question comes with the context on which it was generated, which was included into the prompt for generation, along with instructions and a one-shot prompt for answer formatting purposes, the details of which can be found in Appendix \ref{sec:appendix-a}. The metadata of the sample set can be found in Table~\ref{tab:finsyn_summary_by_difficulty}. 

\textbf{Configurations.} We experiment with 7 models, 4 proprietary and 3 open-source, as detailed in table \ref{tab:elo_ratings}. We employ the following configuration for the models included in our test:

We use a vLLM\cite{kwon2023efficient} backend for serving the open-sourced models (including the Qwen, and the Llama families). We use a temperature of 0.05, with a token limit of 16,384 or 8,192 as determined by the underlying LLM\footnote{in our tested models, only Meta-Llama-3-70B-Instruct uses an 8,192 limit.} For proprietary models, we use their official API services with a temperature of 0.05, with their maximum input token limit provided respectively. We truncate the input exceeding the input limits, which accounts for only <2\% of all samples.

\textbf{Evaluation Framework.} In traditional question-answering (QA) tasks, a range of rule-based and word-matching evaluation metrics have been commonly used to assess model performance. Some of the most prevalent metrics include Exact Match (EM), F1 Score, BLEU (Bilingual Evaluation Understudy) 
 \cite{papineni_bleu_2002}, ROUGE (Recall-Oriented Understudy for Gisting Evaluation) \cite{lin_rouge_2004}, METEOR (Metric for Evaluation of Translation with Explicit ORdering) \cite{banerjee_meteor_2005}, and BERTScore \cite{zhang_bertscore_2020}.

Despite their widespread usage, these metrics exhibit significant limitations, particularly when applied to analytical or open-ended questions. 
 Therefore, in addition to word-matching based metrics, we employ a machine judge for a more interpretable and flexible judging of long-form answers. 
 Specifically, for analytics tasks, we compare each model's response to a baseline (in our case, gpt-4o using text input, with a 1000 base score), ask the judging model (gpt-4-turbo) to reason about which model is more accurate, logical, and clear in its output. The prompts we used are detailed in Appendix \ref{sec:appendix-a}.
 For extraction and calculation tasks, we also provide   accuracy as a metric in Appendix \ref{sec:appendix-c}.


\section{Results}

Table \ref{tab:elo_ratings} demonstrate model performance using ELO scores, with shades of blue and red demonstrating better and worse scoring.

\textbf{Closed-source models like gpt-4-turbo and claude-3-5-sonnet clearly outperform open-source models}, particularly on harder tasks, as evidenced by their significantly higher ELO ratings. GPT-4-turbo leads with an overall ELO rating of 1164.35, excelling in medium difficulty tasks while maintaining strong performance in easier tasks. Claude-3-5-sonnet also performs well across all difficulty levels, showing a more balanced profile with a particularly strong performance in easy tasks. We observe that among open-source model, However, Meta-Llama-3.1-70B-Instruct, an open-source model, shows promise by delivering competitive performance in medium tasks (ELO 1008.80) and a decent overall ELO rating (869.27), standing out as a notable open-source option despite trailing behind the closed-source leaders. We present a more in-depth analysis on what causes the performance gap in Appendix \ref{sec:appendix-b} with the statistics in Appendix \ref{sec:appendix-c}.

\textbf{Models can excel at easier tasks, but be significantly weaker at hard tasks.} Notably, models like gemini-1.5-pro and gemini-1.5-flash excel in simpler tasks, with Easy task ELO ratings reaching up to 967.86 and 880.19 respectively, but struggle with harder tasks, as indicated by their lower Hard ELO ratings (both at 779.98). Similarly, Meta-Llama-3.1-70B-Instruct demonstrates moderate performance in Easy and Medium tasks (ELO 880.19 and 1008.80 respectively) but falls significantly short in Hard tasks, with an ELO rating of 703.98. These patterns highlight the specialization of some models for simple tasks while exposing weaknesses in their ability to handle more complex scenarios. As a side note, we found certain multi-modal models exhibit this pattern as well, which we demonstrate in the complete evaluation results in Appendix \ref{sec:appendix-c}, but we call for future work for further investigation in multi-modal table comprehension benchmarking.

\begin{table}[h]
\centering
\begin{tabular}{|l|c|c|c|c|}
\hline
Model & Overall ELO & Easy ELO & Medium ELO & Hard ELO \\
\hline
gpt-4-turbo & \cellcolor[HTML]{B2D9FF} 1164.35 & \cellcolor[HTML]{99C7FF} 1104.56 & \cellcolor[HTML]{66A3FF} 1372.43 & \cellcolor[HTML]{A0CCFF} 1045.06 \\
claude-3-5-sonnet-20240620 & \cellcolor[HTML]{A0CCFF} 1081.39 & \cellcolor[HTML]{66A3FF} 1250.12 & \cellcolor[HTML]{70B3FF} 1219.87 & \cellcolor[HTML]{FF9D9D} 832.23 \\
gemini-1.5-pro & \cellcolor[HTML]{FFB2B2} 896.81 & \cellcolor[HTML]{FFA8A8} 967.86 & \cellcolor[HTML]{FF9999} 930.28 & \cellcolor[HTML]{FF9C9C} 779.98 \\
gemini-1.5-flash & \cellcolor[HTML]{FFADAD} 880.88 & \cellcolor[HTML]{FF8080} 880.19 & \cellcolor[HTML]{FFAAAA} 969.48 & \cellcolor[HTML]{FF9C9C} 779.98 \\
Meta-Llama-3.1-70B-Instruct & \cellcolor[HTML]{FFB5B5} 869.27 & \cellcolor[HTML]{FF8080} 880.19 & \cellcolor[HTML]{FFB7B7} 1008.80 & \cellcolor[HTML]{FF7373} 703.98 \\
Qwen2-72B-Instruct & \cellcolor[HTML]{FFB6B6} 867.81 & \cellcolor[HTML]{FF8181} 884.54 & \cellcolor[HTML]{FF8A8A} 912.63 & \cellcolor[HTML]{FFB1B1} 793.03 \\
Meta-Llama-3-70B-Instruct & \cellcolor[HTML]{FF5151} 772.05 & \cellcolor[HTML]{FF5A5A} 853.85 & \cellcolor[HTML]{FF6868} 833.20 & \cellcolor[HTML]{FF0000} 606.52 \\
\hline
\end{tabular}
\vspace{2mm}
\caption{ELO ratings across difficulties. Baseline is GPT4o, and we start with a base score of 1000.}
\label{tab:elo_ratings}
\end{table}

\vspace{-5mm}

\section{Conclusion}
In this paper, we introduce TableQuest, a new benchmark, that addresses the gaps by evaluating LLMs' ability to reason with tables in natural, table-rich context of financial reports. Our results with 7 state-of-the-art models reveal that, while LLMs can locate facts with reasonable accuracy, they struggle with complex reasoning and multi-step calculations. This highlights the need for future improvements in LLMs to enhance their proficiency in tasks requiring comprehensive table comprehension.










\bibliographystyle{unsrt}
\bibliography{references,custom}

\clearpage
\appendix

\section{Prompts}
\label{sec:appendix-a}

Below is the inference prompt which we use on the 7 models tested.

\begin{tcolorbox}[colback=white]
\small

\textbf{[System]}\\

You are a helpful question-answering chatbot.\\

\textbf{[User]}\\

You are asked to answer questions based on provided material containing tables and text in HTML. You should layout your reasoning steps (if necessary) to obtain the correct answer (marked by "steps"), then provide a succinct answer (marked by "answer") in JSON. If units can be known, the answer should include units such as \$, \%, million and etc. \\

\textcolor{blue}{<one-shot example>}\\

\#\# Context: \\

\textcolor{blue}{<context>}\\

\#\# Question: \\

\textcolor{blue}{<question>}\\

\label{text:judger}
\end{tcolorbox}

Below is the prompt we use for our judging model, gpt-4-turbo.

\begin{tcolorbox}[colback=white]
\small
\textbf{[User]}\\

You will be presented with a question, the context, a reference answer marked with "gt", and two model predictions marked by "pred\_1" and "pred\_2" in JSON format. Compare the two predictions and evaluate which one better answers the question based on the reference answer. Output your rationale (marked by "rationale" in JSON) for your decision, and then your decision (marked by "better\_model" in JSON, with possible values "model\_1", "model\_2", or "tie") in JSON format.

Please use the following listed aspects and their descriptions as evaluation criteria:
\begin{itemize}
    \item \textbf{Accuracy}: The assistant's answer is semantically consistent with the gold answer, preferably with the reference answer verbatim;
    \item \textbf{Clarity}: The assistant's answer is clear and easy to understand;
    \item \textbf{Logical}: The assistant's reasoning steps are clear and reasonable, does not make gigantic leaps between steps and makes assumptions without clear evidence;
\end{itemize}

\#\# Context in JSON:\\
\{
    \textbf{question}: \textcolor{blue}{<question>}\\
    \textbf{gt}: \textcolor{blue}{<answer>}\\
    \textbf{pred\_1}: \textcolor{blue}{<LLM1's response>}\\
    \textbf{pred\_2}: \textcolor{blue}{<LLM2's response>}\\
    ...
\}

\#\# Output:
\end{tcolorbox}

\label{paper:GPT4-as-the-Judge}






Below is the synthesis prompts we used for our multi-round data generation. Each generated question-answer pair from a lower difficulty are passed into the prompt serving as a negative example, encouraging the question-answer pair to be accurate to the intended difficulty level.

\begin{tcolorbox}[colback=gray!5!white, colframe=gray!75!black, title=Prompt 1]
You are an underwriter examining a company's performance to make a critical decision which has a critical aspect to your company and your job. Generate a question that you have for the snippet annual report from \{company\} \{year\}. Then, present your answer.
\end{tcolorbox}

\vspace{0.5cm}

\begin{tcolorbox}[colback=gray!5!white, colframe=gray!75!black, title=Prompt 2]
You are designing a test question for middle schoolers' ability to interpret information from the table. Generate a question that asks for the information from a cell. Then provide the correct answer.

...

\textbf{Output:}
\end{tcolorbox}

\vspace{0.5cm}

\begin{tcolorbox}[colback=gray!5!white, colframe=gray!75!black, title=Prompt 3]
You are now designing a test question for high schoolers' ability to calculate based on given instructions. Generate a question which involves numbers not directly mentioned in the table. Then, provide the steps to calculate. Finally, present the correct answer (a number or a few words, NOT a sentence).

...

\textbf{Output:}
\end{tcolorbox}

\vspace{0.5cm}

\begin{tcolorbox}[colback=gray!5!white, colframe=gray!75!black, title=Prompt 4]
You're an experienced university professor designing test questions for corporate finance. Based on the material provided, generate an analytical question that focuses on a fraction of the content. Then, provide a detailed step-by-step reference reasoning steps. Finally, provide a concise reference answer which should be less than 3 sentences long for the ease of grading. 

...

\textbf{Output:}
\end{tcolorbox}

\section{Quality assurance on the data synthesis pipeline}
\label{sec:appendix-a.2}

Our process involved generating an initial set of questions, followed by iterative feedback sessions with these experts to refine and enhance the quality of the questions. To ensure a thorough evaluation, we selected a representative sample of 50 questions from each difficulty level for human annotation. For this task, we engaged four STEM graduates who were responsible for locating the sources of answers, verifying the alignment of human-provided answers with the synthetic ones, and assessing whether the quality of the questions appropriately matched their designated difficulty levels. Throughout this iterative process, we maintained a high standard of accuracy. We identified the most common error cases, which were then used as in-context examples to further fine-tune the performance of GPT-4-Turbo, ensuring it could accurately re-evaluate the question-answer pairs. This approach allowed us to systematically address and rectify discrepancies, enhancing the overall robustness and validity of our benchmark. Finally, the refined set of question-answer pairs demonstrated high fidelity to real-world scenarios, reflecting the domain knowledge and quality standards initially set out.


\section{Case Study}
\label{sec:appendix-b}



Given the gap in open-source/proprietary model performance, we conduct a qualitative analysis with one model from each world as representatives, llama-3.1-70B-Instruct and GPT-4-Turbo. We employ the most recent reasoning models to date, o1-preview, to judge and critique both responses on the same question for comparison. On a 3-point scale scoring system, GPT-4-Turbo leads by 0.34 on average. We identified 12.5\% of the evaluation samples on which GPT-4-Turbo scored 2 or higher than Llama-3.1-70B-Instruct.

The performance disparity between Llama and GPT-4-Turbo in the financial analysis benchmark reveals limitations in Llama's capabilities. Firstly, the prevalence of empty responses, particularly in questions requiring straightforward fact retrieval, suggests deficiencies in Llama's long-context information retrieval and instruction following output generation mechanisms. 
Secondly, the cases of calculation errors and misinterpretation of numerical data, indicates weaknesses in its mathematical reasoning abilities and financial data comprehension. These errors ranged from failing to calculate specific percentages to misinterpreting financial trends, highlighting a need for enhanced numerical processing capabilities. 
Lastly, Llama demonstrated difficulties with complex questions requiring multi-step reasoning. In these instances, Llama often provided incomplete analyses or failed to address key aspects of multi-faceted queries, suggesting limitations in its ability to decompose and address complex problems. These findings collectively point to areas of improvement for all open sourced models, particularly in domains requiring precise numerical analysis and multi-step reasoning. Below we showcase five representative examples of the differences in the judgement from the judge to the two models.

\subsection{Case 1: Empty Response vs Detailed Answer (ID: 93)}

\begin{tcolorbox}[title=Llama Response, colback=red!5!white, colframe=red!75!black]
The model prediction is completely empty, providing no answer to the question asked.
\end{tcolorbox}

\begin{tcolorbox}[title=GPT-4-Turbo Response, colback=green!5!white, colframe=green!75!black]
The prediction accurately identifies the number of units in the property 'The Grand' as 243, which matches the reference answer verbatim. The prediction also provides a clear and logical step-by-step explanation of how the answer was derived, including locating the property in a table and identifying the relevant column for unit numbers.
\end{tcolorbox}

\subsection{Case 2: Calculation Error vs Accurate Computation (ID: 186)}

\begin{tcolorbox}[title=Llama Response, colback=red!5!white, colframe=red!75!black]
The reference answer states a specific increase of approximately 1115.38\% in non-subscription revenue due to the merger. However, the model prediction does not mention this percentage increase and instead concludes that the exact impact cannot be quantified, which contradicts the reference answer.
\end{tcolorbox}

\begin{tcolorbox}[title=GPT-4-Turbo Response, colback=green!5!white, colframe=green!75!black]
The prediction provides a detailed and logical step-by-step analysis of the non-subscription revenue growth from 2021 to 2023, including calculations of the absolute and percentage increase. It also correctly identifies the impact of the IHS Markit merger as a substantial positive influence on this growth, which aligns with the reference answer.
\end{tcolorbox}




\subsection{Case 3: Misinterpretation of Question (ID: 991)}

\begin{figure}[htbp]
\centering
\includegraphics[width=\textwidth]{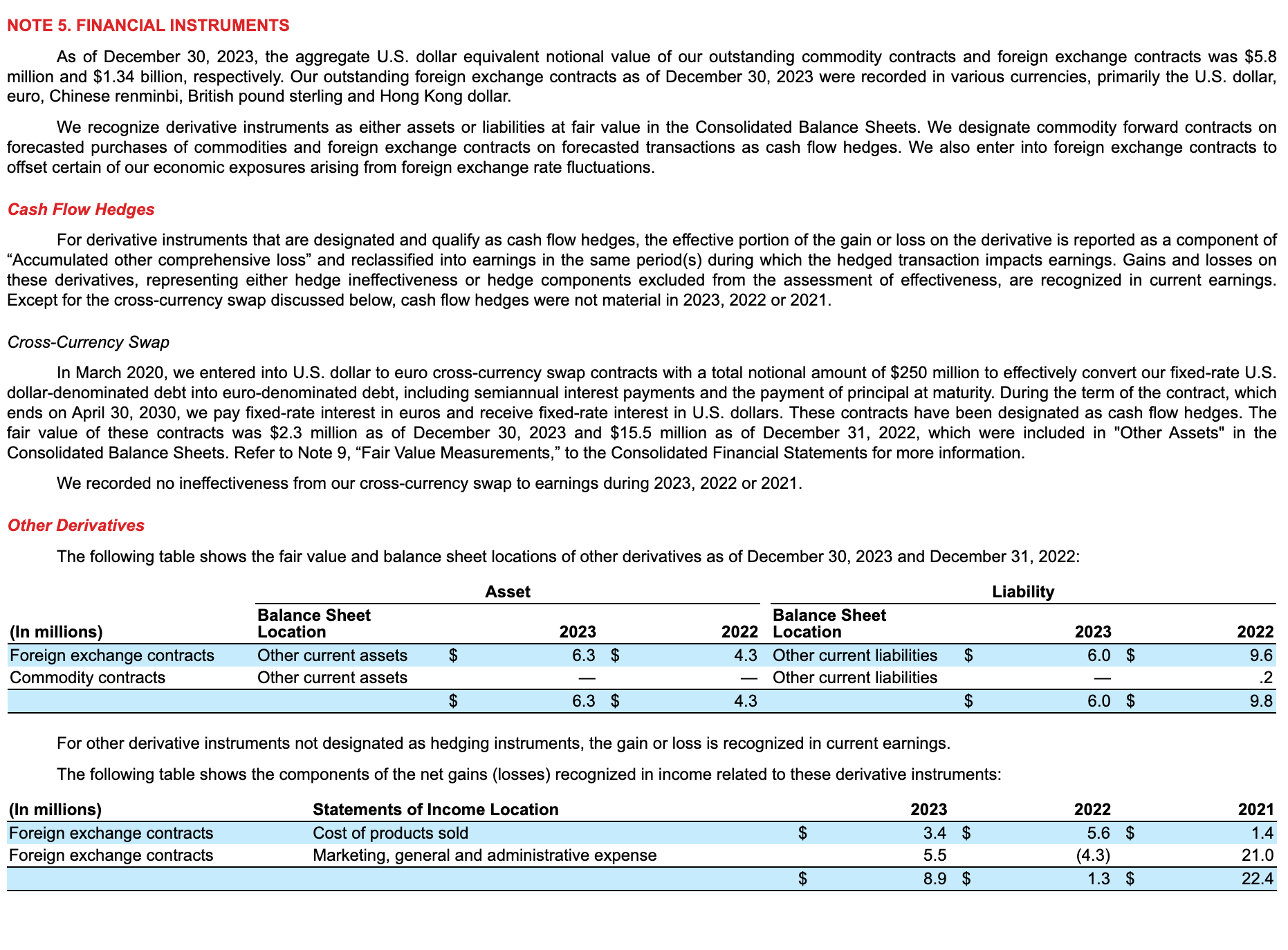}
\caption{Excerpt from the company's financial report}
\label{fig:financial_report}
\end{figure}

Below we have the question, ground truth answer, and the context document in Figure \ref{fig:financial_report}. The model outputs and the judgements are laid out in the following.



\begin{tcolorbox}[title=Question and Ground Truth]
\textbf{Question:} Analyze the effectiveness of the cash flow hedge strategy for the cross-currency swaps by comparing the fair values of the contracts between 2022 and 2023.

\textbf{Ground Truth Answer:} The fair value of the cross-currency swaps decreased by \$13.2 million from 2022 to 2023, indicating reduced effectiveness or negative economic impact on the hedge during this period.
\end{tcolorbox}

\vspace{1em}

\begin{minipage}[t]{0.49\textwidth}
\begin{tcolorbox}[colback=red!5!white,colframe=red!75!black,title=LLaMA Prediction,fontupper=\footnotesize,equal height group=prediction]
\texttt{
\{ 
  "steps": "To analyze the effectiveness of the cash flow hedge strategy for the cross-currency swaps, we need to calculate the change in fair value of the contracts between 2022 and 2023. The change in fair value is calculated as the difference between the fair value in 2023 and the fair value in 2022. Then, we calculate the percentage change in fair value by dividing the change in fair value by the original fair value in 2022 and multiplying by 100.", 
  "answer": \{ 
    "change\_in\_fair\_value": "-\$13.2 million", 
    "percentage\_change\_in\_fair\_value": "-85.16\%" 
  \} 
\}
}
\end{tcolorbox}
\end{minipage}
\hfill
\begin{minipage}[t]{0.49\textwidth}
\begin{tcolorbox}[colback=green!5!white,colframe=green!75!black,title=GPT Prediction,fontupper=\footnotesize,equal height group=prediction]
\texttt{
\{ 
  "2023": "\$2.3 million", 
  "2022": "\$15.5 million", 
  "Analysis": "The fair value of the cross-currency swaps decreased from \$15.5 million in 2022 to \$2.3 million in 2023, indicating a significant reduction in the value of these hedges over the year. This could suggest a decrease in effectiveness or adverse market conditions impacting the value of these financial instruments." 
\}
}
\end{tcolorbox}
\end{minipage}

\vspace{1em}

\begin{minipage}[t]{0.49\textwidth}
\begin{tcolorbox}[colback=red!5!white,colframe=red!75!black,title=LLaMA Analysis,fontupper=\footnotesize,equal height group=analysis]
\begin{itemize}
    \item Quantitative approach with specific calculations
    \item Calculated change in fair value: -\$13.2 million
    \item Calculated percentage change: -85.16\%
    \item Interpreted decrease as indicating effectiveness in reducing foreign exchange rate exposure
    \item Provided a step-by-step analysis with precise calculations
    \item Misinterpreted the decrease as a positive outcome
    \item Lacked broader context in the analysis
\end{itemize}
\end{tcolorbox}
\end{minipage}
\hfill
\begin{minipage}[t]{0.49\textwidth}
\begin{tcolorbox}[colback=green!5!white,colframe=green!75!black,title=GPT Analysis,fontupper=\footnotesize,equal height group=analysis]
\begin{itemize}
    \item Qualitative approach focusing on overall trends
    \item Did not provide specific calculations
    \item Interpreted decrease in fair value as suggesting reduced effectiveness or adverse market conditions
    \item Offered a concise summary highlighting potential implications
    \item Contextualized the change within broader market conditions
    \item Lacked detailed quantitative analysis
    \item Provided a more accurate assessment of the hedge's effectiveness
\end{itemize}
\end{tcolorbox}
\end{minipage}





\section{Complete Result}
\label{sec:appendix-c}

We layout the complete evaluation statistics of the previous 7 models, and 2 multi-modal variants using images as the context (gpt-4o, and gemini-1.5-pro) under different evaluation frameworks in Table \ref{tab:model_comparison}.
 
\rowcolors{2}{gray!15}{white}
\begin{table}[b]
\renewcommand{\arraystretch}{1.5}
\centering
\begin{tabular}{|l|c|c|c|c|c|}
\hline
{Model} & \multicolumn{2}{c|}{ELO Rating} & \multicolumn{3}{c|}{Accuracy (\%)} \\
\cline{2-6}
 & Overall & Hard & Overall & Easy & Medium \\
\hline
gpt-4o-vision & 892.46 & 797.37 & 70.0 & 72.5 & 67.5 \\
gemini-1.5-pro & 896.81 & 779.98 & 66.25 & 70.0 & 62.5 \\
gpt-4-turbo & 1164.35 & 1045.06 & 65.625 & 63.75 & 67.5 \\
claude-3-5-sonnet-20240620 & 1081.39 & 832.23 & 65.625 & 67.5 & 63.75 \\
Qwen2-72B-Instruct & 867.81 & 793.03 & 65.0 & 65.0 & 65.0 \\
gemini-1.5-flash & 880.88 & 779.98 & 64.375 & 68.75 & 60.0 \\
gemini-1.5-pro-vision & 760.52 & 639.71 & 54.375 & 62.5 & 46.25 \\
Meta-Llama-3-70B-Instruct & 772.05 & 606.52 & 51.875 & 53.75 & 50.0 \\
Meta-Llama-3.1-70B-Instruct & 869.27 & 703.98 & 48.75 & 48.75 & 48.75 \\
\hline
\end{tabular}
\vspace{5mm}
\caption{Model Performance Comparison: ELO Rating and Accuracy}
\label{tab:model_comparison}
\end{table}






\section{Dataset Details}
\label{sec:appendix-e}

\begin{figure}
    \centering
    \includegraphics[width=1\linewidth]{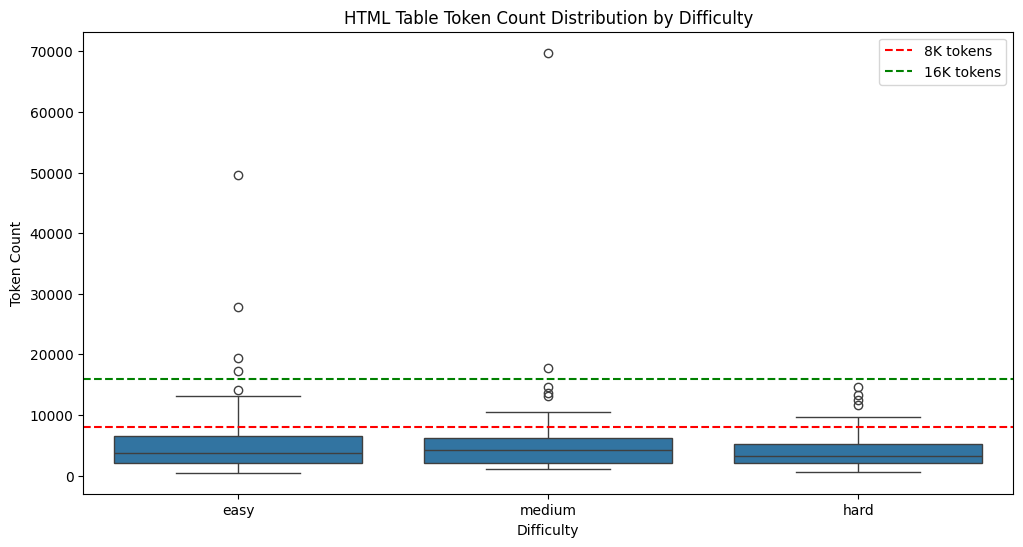}
    \caption{Input lengths statistics.}
    \label{fig:input-length}
\end{figure}

We detail the input length by token count in Figure \ref{fig:input-length}. As demonstrated by the green line, the inputs exceeding the 16k token limits are the outliers. Given that all but one model tested in our experiments accept 16k tokens, we believe that the truncation employed to fit context limit does not pose a significant disadvantage to open-source models with a smaller context limit.


\begin{table}[h]
\centering
\begin{tabular}{lccc}
\hline
Length & Distribution (\%) & Recall \\
\hline
0-5 & 26.7\% & 0.704  \\
5-10 & 15.0\% & 0.788  \\
10-15 & 19.3\% & 0.816  \\
15-20 & 15.8\% & 0.782  \\
20+ & 23.2\% & 0.642  \\
Avg. & 100\% & 0.736 \\
\hline
\end{tabular}
\vspace{5mm}
\caption{Metrics by Answer Length. The recall slump within the short-form answer group (0-5 tokens) is primarily attributable to the slight differences in calculated and claimed number strings.}
\
\label{tab:metrics-by-length}
\end{table}

\section{Self-supervised question-answer synthesis validation}
\label{sec:appendix-f}

Despite extensive efforts from both human and machine annotators, the complexity and the domain knowledge required to validate whether the questions are meaningful, or whether the answer aptly responds to the question, i.e. whether our external supervision setup was, in fact, factual - remains a challenging point to prove. Therefore, in an attempt to verify the efficacy of the data synthesis pipeline, we adapted it for a new domain, diversifying our data sources. More importantly, this adaptation enabled us to leverage self-supervision signals within the context document to fact-check synthesized answers reliably.

Specifically, we selected recent peer-reviewed papers from reputable artificial intelligence conferences as source material for our data synthesis pipeline. Our selection was based on three reasons: (1) Conference papers in this domain often contain rich tabular content, which is frequently referenced in the main text with detailed analysis; (2) The expert peer-review process ensures a certain quality of content; and (3) The AI field has a high volume of contributions, providing a solid foundation for scaled analysis.

We collected 936 academic papers from January 2022 to May 2024, utilizing a subset of a recently released dataset \cite{minzheng_wang_mozerwangloong_2024} focused on long-context reasoning. Since the data was obtained in markdown format through OCR from the original PDFs, it was incompatible with our existing synthesis pipeline, which was designed for HTML-format handling. We employed several preprocessing steps: (a) splitting each markdown paper into chunks according to its native formatting and extracting tables and headers using regular expressions; (b) gathering all chunks that explicitly reference each table to serve as context for question synthesis; and (c) converting all markdown tables into canonical HTML form using the \textit{markdownify} package. These steps transformed the original data source into a format compatible with our synthesis pipeline and provided high-quality evidence for verifying synthetic answers. All other configurations, including model settings and prompts, were kept consistent with the original setup to preserve the validity of the validation process. Synthetic questions are generated with a requirement to cite a grounding sentence from the evidence pool, while synthetic answers are created based solely on the corresponding tables and questions, without referencing direct evidence from the paper. Following quality filtering procedures, we obtained 2,100 question-answer pairs, with 700 representing each difficulty level.

To evaluate the faithfulness of synthetic answers to the source, we used recall as our evaluation metric. Specifically, we leveraged the \textit{nltk} package to tokenize each answer string and the corresponding evidence string, which theoretically contains the answer. We then identified the common tokens between the two and calculated the recall rate based on the total tokens in the answer string. The results (as shown in Table \ref{tab:metrics-by-length}) indicate an average recall rate of 0.74, demonstrating our synthesis model's (GPT-4-turbo) strong capability to produce accurate answers that align with the source.

In conclusion, our adaptable data synthesis pipeline demonstrates robust performance in generating factually grounded question-answer pairs, particularly through its effective handling of domain-specific content like academic papers with tabular data leveraging a self-supervision signal. This versatility highlights its potential for applications across diverse domains that require structured and unstructured data synthesis, making it a valuable tool for generating and verifying knowledge in areas such as long-context reasoning and fact-checking. Future extensions of this pipeline could further enhance its applicability by incorporating additional domains and refining its adaptability to varying data formats.

\end{document}